\documentclass[letterpaper]{article}
\usepackage{iccc}
\usepackage{times}
\usepackage{helvet}
\usepackage{courier}
\usepackage{xcolor}
\usepackage{url}

\usepackage{graphicx}
\usepackage{amsmath,amssymb} 
\usepackage{color}
\usepackage{epsfig}
\usepackage{pdfpages}
\usepackage{caption}
\usepackage{subcaption}

\usepackage{multirow}
\usepackage{rotating}
\usepackage{booktabs}
\usepackage{algpseudocode}
\usepackage{algorithm}
\usepackage{etoolbox}

\newcommand{\arxiv}[1]{\textcolor{black}{#1}}

\newtoggle{arxiv} 
\toggletrue{arxiv} 

\title{Exploring Crowd Co-creation Scenarios for Sketches}

\iftoggle{arxiv}{
\author{Devi Parikh$^{1,2}$ and C. Lawrence Zitnick$^{1}$\\
$^{1}$Facebook AI Research \\
$^{2}$Georgia Tech\\
parikh@gatech.edu \quad zitnick@fb.com\\
}
}{
\author{Anonymous Authors\\
Anonymous Institution \\
anonymous@email.an\\
}
} 

\begin{document} 
\maketitle


\begin{abstract}
\begin{quote}
As a first step towards studying the ability of human crowds and machines to effectively co-create, we explore several human-only collaborative co-creation scenarios. The goal in each scenario is to create a digital sketch using a simple web interface. We find that settings in which multiple humans iteratively add strokes and vote on the best additions result in the sketches with highest perceived creativity (value + novelty). Lack of collaboration leads to a higher variance in quality and lower novelty or surprise. Collaboration without voting leads to high novelty but low quality.

\end{quote}
\end{abstract}


\section{Introduction}
How can one best collaborate with humans in a creative process?
Insights towards this can inform what roles machines can (or should not) play when co-creating with humans. 

\begin{figure}[t]
\centering
\includegraphics[width=0.6\columnwidth]{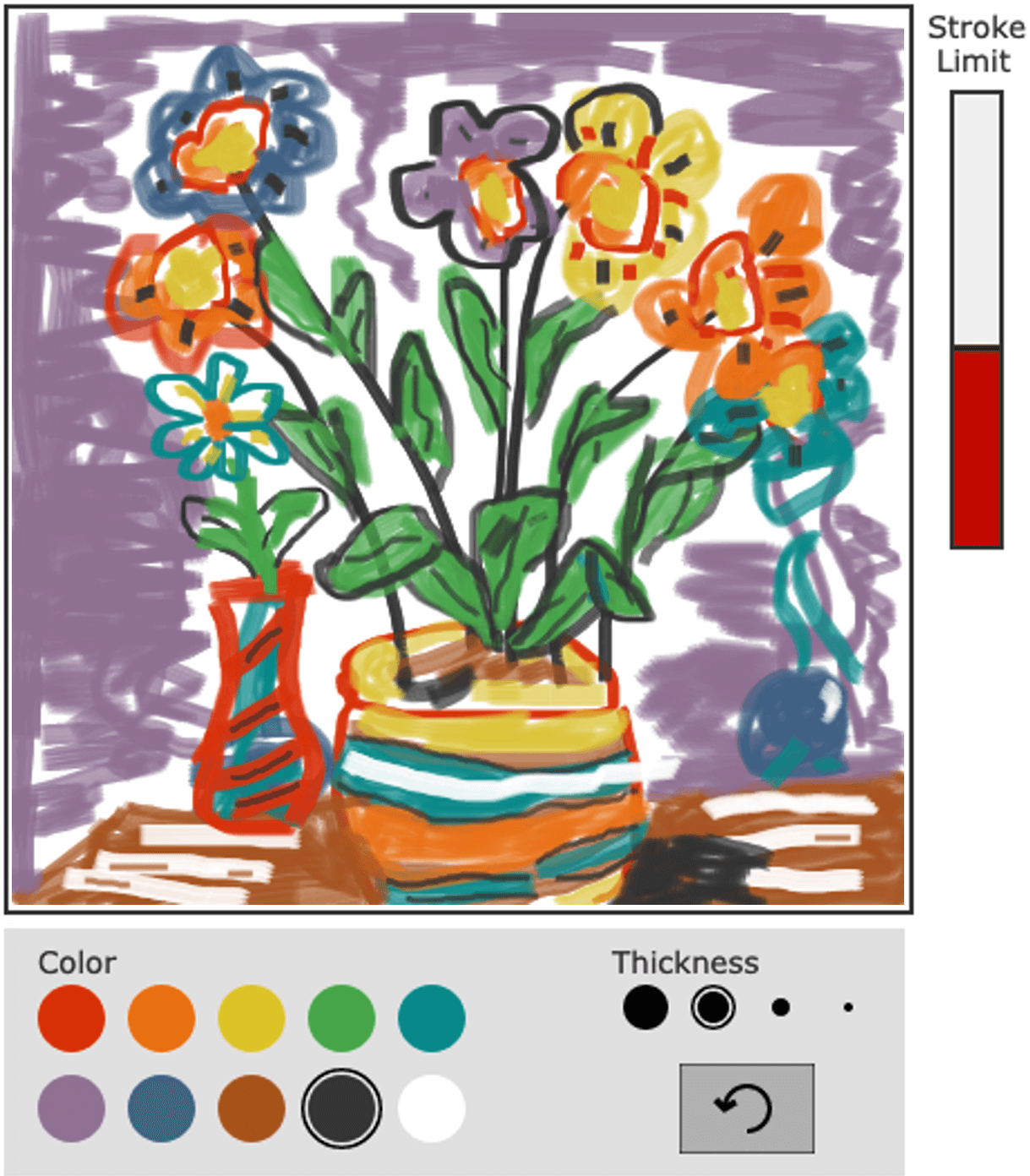}
\caption{As a first step towards human-machine  co-creation, we explore human-human collaboration for creating digital sketches on a simple web interface shown above. Video:  
\url{https://youtu.be/9fikuKPYPd0}
}
\label{fig:interface}
\vspace{-10pt}
\end{figure}

Specifically, we consider a scenario where agents take turns collaboratively drawing a sketch on a simple web interface (Figure~\ref{fig:interface}). During each iteration, multiple agents propose strokes to add to the sketch. 
Agents then vote on the proposals, and the preferred set of strokes is added to the sketch. This process is repeated 
for a fixed number of iterations to create a final sketch.

The roles of creating stroke proposals and voting could each be fulfilled by either humans (H) or machines (M). Borrowing terminology from Generative Adversarial Networks \cite{gans}, we can call the former role a generator (G), and the latter a discriminator (D). This allows for 4 \{H,M\} $\times$ \{G,D\} co-creation scenarios. Further, different individuals could play the role of generators/discriminators across iterations, leading to \emph{crowd} co-creation. 

\begin{figure*}[t!]
    \centering
     \includegraphics[width=0.9\textwidth]{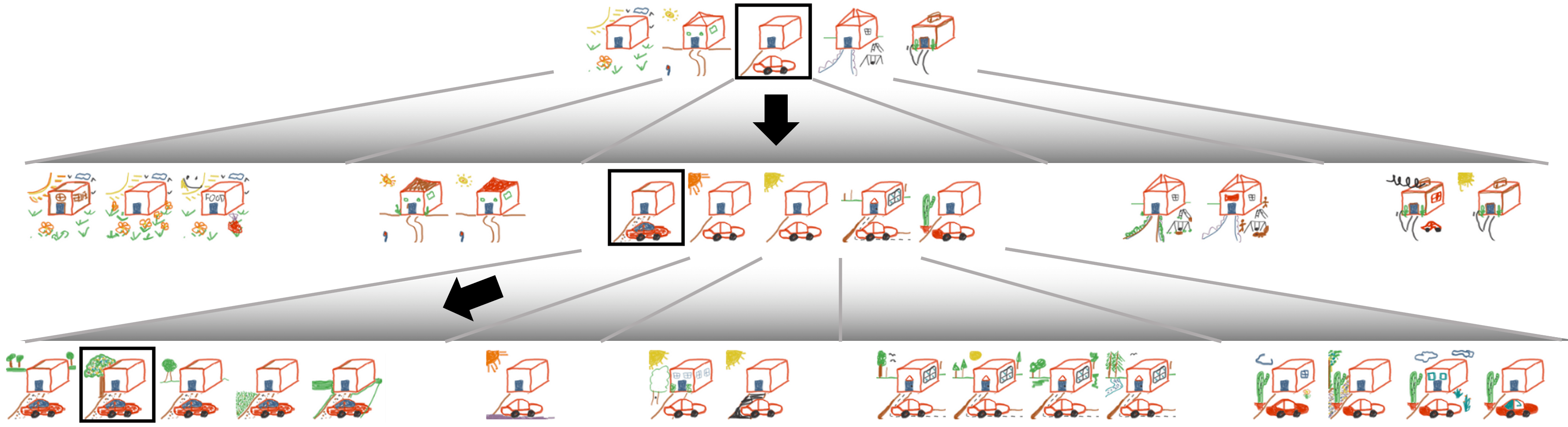}
    \caption{Few iterations of a sketch being created in the \textbf{Collaborative + voting} scenario. Once a parent sketch gets five children, it gets selected as the next iteration of the sketch (black outline), and the five children become the parents for the next iteration. Temporal visualization: \url{https://youtu.be/JQmGALAhhMU}.
    \iftoggle{arxiv}{\arxiv{Examples with all iterations: Figures~\ref{fig:whole_tree_6} and \ref{fig:whole_tree_17}.}}{}
    }
    \label{fig:tree}
    \vspace{-10pt}
\end{figure*}

In this work, as a step towards human-machine co-creation, we study various \emph{human-human} crowd co-creation scenarios. In the first, \textbf{Individual}, a single human creates the entire sketch (no discriminator D, and no crowd). Second, in \textbf{Collaborative} the sketch is generated by multiple human agents (crowd) iteratively taking turns adding strokes. That is, all the agents act as generators G and there is no voting or discriminator D. The third, \textbf{Collaborative + voting}, is where multiple human agents (generators) propose new strokes at each iteration. Another set of human agents (discriminators) vote on which set of strokes to add to the sketch. Finally, we explore \textbf{Individual with collaborative prompts}, for which the crowd is involved indirectly. A single human creates the entire sketch, but by following text prompts that describe the evolution of a sketch that was created in the \textbf{Collaborative} scenario.

We evaluate the qualitative difference between the sketches produced via these four scenarios. We find that the collaborative setting \emph{with} a voting mechanism (\textbf{Collaborative + voting}) leads to sketches that are rated by human subjects as most creative (and are preferred along a variety of other dimensions). The lack of either one of these components results in less creative sketches: \textbf{Individual} sketches have decent quality (value) but low novelty, while \textbf{Collaborative} sketches have high novelty but low value. \textbf{Individual with collaborative prompts} results in high novelty but even worse quality. Overall, among these four scenarios, \textbf{Collaborative + voting} best hits the sweet spot for creativity: value + novelty~\cite{boden92}.


\section{Related Work}

In the context of crowd-based sketching, \cite{Tuite2012} analyze user actions and large-scale behavior patterns in 50k sketches from Sketch-a-bit, a collaborative mobile drawing application. Different from our incremental contribution + voting mechanisms, \cite{Yu2011} and \cite{Gingold2012} explore combination and averaging of sketches respectively as collaboration strategies.

\begin{figure*}[t]
    \centering
    \includegraphics[width=0.9\textwidth]{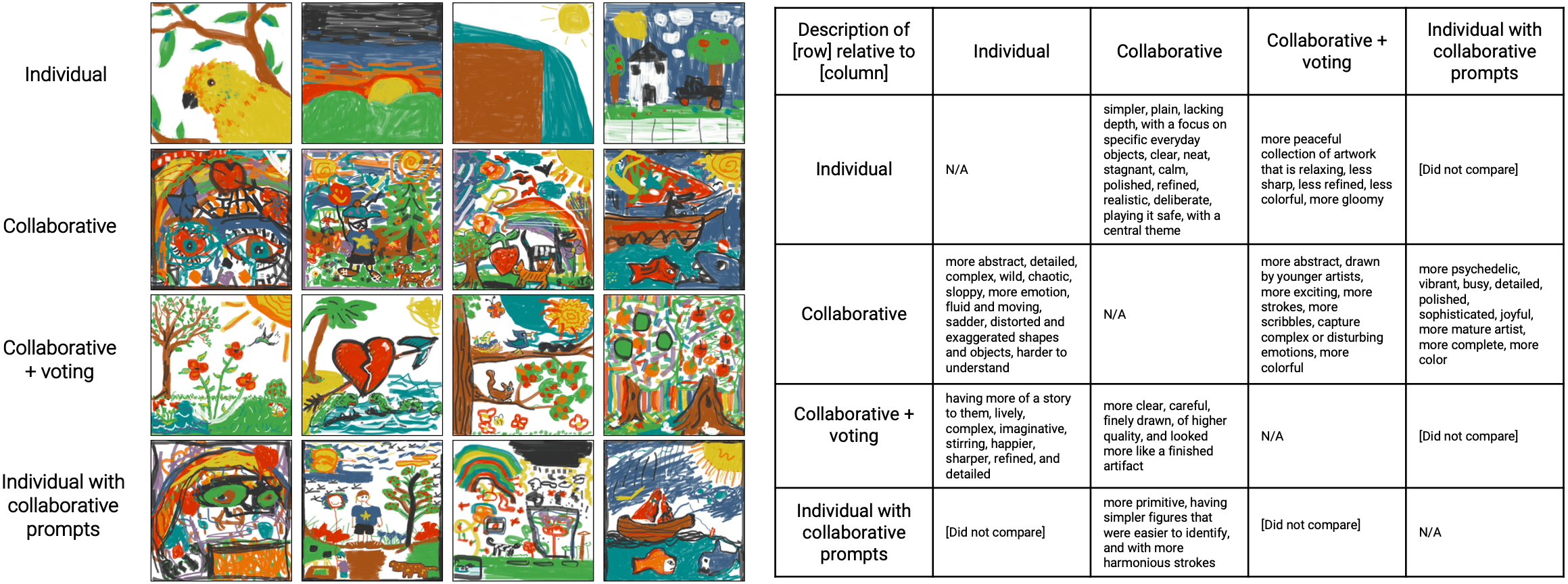}
    \caption{Example sketches from four co-creation scenarios along with differences identified by human subjects between sketches from pairs of scenarios. \textbf{Collaborative + voting} involves $\sim$12.5 times the individuals, and so was run for 20 instead of 30 iterations. For comparison, \textbf{Collaborative} sketches are also shown at 20 iterations. \iftoggle{arxiv}{\arxiv{More sketches from the four scenarios can be seen in Figures~\ref{fig:all_ind}, \ref{fig:all_col}, \ref{fig:all_col_vot}, and \ref{fig:all_ind_prmpt} respectively.}}{}
    }
    \label{fig:example_sketches}
    \vspace{-10pt}
\end{figure*}

Several AI systems have been trained to recognize sketches (e.g., models trained on The Quick, Draw! Dataset\footnote{{\tiny \url{https://github.com/googlecreativelab/quickdraw-dataset}}}). These may form useful building blocks for the next stages of our work. However, as seen in Figure~\ref{fig:example_sketches}, our sketches tend to be complex scenes and often abstract as opposed to concrete individual objects, which has been the focus of most existing work in automatic sketch recognition. There is also work on  generating images based on sketches~\cite{sketchyGANS2018}.

\cite{Davis2016} employ a cognitive science framework called participatory sense-making to study co-creation in sketches. Central to their study is the back and forth interaction (dialog) between the human and machine as they take turns. Our work is focussed on a crowd setting where no two agents interact again in the future. \cite{Karimi2020} study human-AI co-creativity in the context of humans sketching for a particular design goal. Our work falls in the category of ``casual creators''~\cite{Compton2015} -- systems that support exploratory as opposed to goal-driven creativity. 


\section{Sketching Interface}

Human agents create sketches using the JavaScript based interface shown in Figure~\ref{fig:interface}. Strokes can be varied across four thicknesses and ten colors and have a paint-like texture. The number and length of the strokes an agent may draw is limited during each iteration. Feedback on how close they are to the cutoff is provided in real-time by the stroke limit bar. Thicker strokes count more towards the limit. Strokes drawn by the agent during the current iteration may be undone. See \url{https://youtu.be/9fikuKPYPd0} for a video of the interface. Our interface is publicly available.


\begin{figure}[t]
    \centering
    \includegraphics[width=\columnwidth]{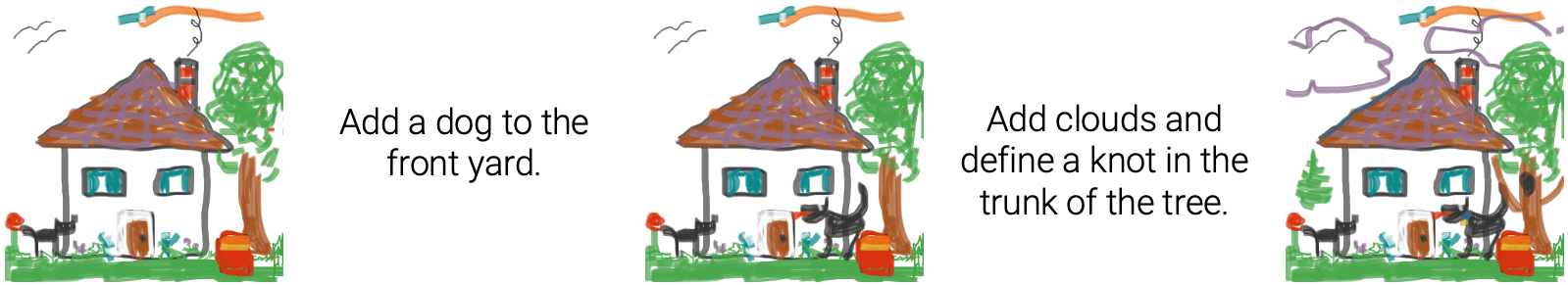}
    \caption{Example prompts used in the \textbf{Individual with collaborative prompts} scenario.}
    \label{fig:example_prompts}
\end{figure}

\begin{figure}[t]
    \centering
    \includegraphics[width=\columnwidth]{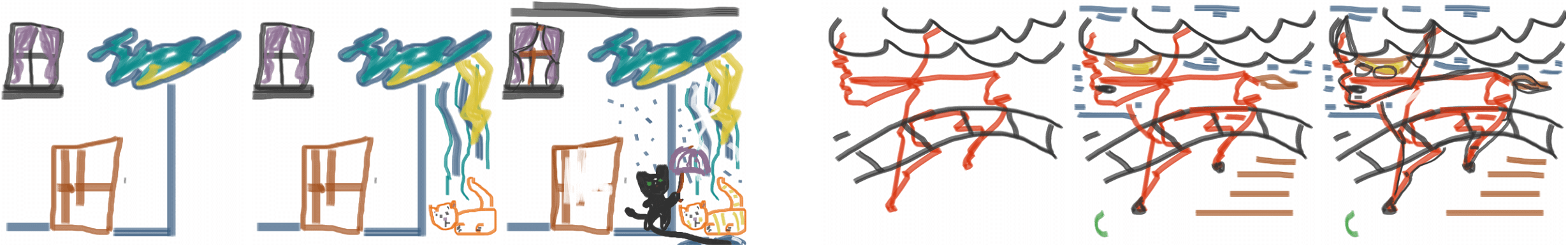}
    \caption{Evolution of example sketches in the \textbf{Collaborative} scenario. Left: Focus of the sketch shifts from the house to the cat in the rain outside the house. Right: Faced with seemingly incoherent strokes, subjects emphasize structure they see in it so subsequent subjects can add to it. \iftoggle{arxiv}{\arxiv{More examples of sketches evolving are in Figures~\ref{fig:evo_col} and \ref{fig:evo_col_vot}.}}{}}
    \label{fig:evolution}
    \vspace{-10pt}
\end{figure}

\section{Co-creation Scenarios}

We explore four scenarios for collaborative human-human sketch co-creation. In every scenario, the sketch starts with a blank canvas. During each iteration, a limited number of strokes may be added. The limit roughly corresponds to five medium-thickness strokes spanning the width of the canvas. 30 iterations are used to create each sketch. Unless stated otherwise, we collected 20 sketches for each scenario. All our studies were conducted on Amazon Mechanical Turk. Subjects can not submit their work till they have contributed the required amount of strokes to the canvas.

\textbf{Individual.} 
The entire sketch is created by a single individual. That is, a single human agent adds all 30 iterations of strokes to ``Create a beautiful, detailed, coherent painting!''.

\textbf{Collaborative.} 
A different human agent contributes strokes for each iteration of the sketch. That is, 30 unique individuals contribute to a sketch. The first subject sees a blank canvas and adds strokes. Every subsequent subject is shown the partial sketch and asked to add to it. They cannot undo strokes from earlier contributors. The prompt is ``Let's collectively create a beautiful, detailed, coherent painting!''. Subjects are given the additional instruction to consider the kind of painting being created and the stage of the painting when deciding upon which strokes to draw.

\textbf{Collaborative + voting.} 
Each subject contributes strokes to a sketch of their choosing from a set of five starting sketch variations. We refer to the chosen starting sketch as a parent, and the sketch created by a subject as the chosen sketch's child. During each iteration, sketches are gathered until a parent is selected five times. Its children then replace the current five parents and the process is repeated. Children of parents selected less than five times are discarded. See Figure~\ref{fig:tree}\iftoggle{arxiv}{\arxiv{, Figures~\ref{fig:whole_tree_6} and \ref{fig:whole_tree_17}}}{} and \url{https://youtu.be/JQmGALAhhMU}\iftoggle{arxiv}{ \arxiv{for more examples}}{}.

This voting strategy allows for the most promising versions of a sketch to go forward. This scenario is robust to the strokes added by any one individual. Of course, it is also significantly more ``expensive''. In the best case scenario where a single parent gets all 5 children and none of the other parents get a child, it takes 5 times the amount of strokes to create a sketch compared to \textbf{Collaborative}. In the worst case, all 5 parents get 4 children each before a parent gets a fifth child. This would result in 21 times the number of strokes. In practice we found this factor to be about 12.5 times. Given the increased cost, we reduced the number of iterations in this scenario to 20 (as opposed to 30). On average, 250 unique individuals contribute to a single sketch.

\textbf{Individual with collaborative prompts.}
A single individual creates an entire sketch using instructive text prompts provided at each iteration. The individual is instructed to follow the prompts when drawing. The text prompts are generated by asking another \text{individual} to describe what changed in a sketch from one iteration to the next in the \textbf{Collaborative} scenario. All text prompts for a sketch are written by a single individual. This is an interesting hybrid of having a single creator, but being guided through prompts that describe the evolution of a sketch as created by 30 unique individuals. We collected three sets of text descriptions for each of the 20 \textbf{Collaborative} sketches. This resulted in a total of 60 \textbf{Individual with collaborative prompts} sketches. In our evaluation, we consider 20 sketches (randomly picking 1 out of the set of 3). See Figure~\ref{fig:example_prompts} for example prompts.


\begin{figure*}[t]
    \centering
    \includegraphics[width=\textwidth]{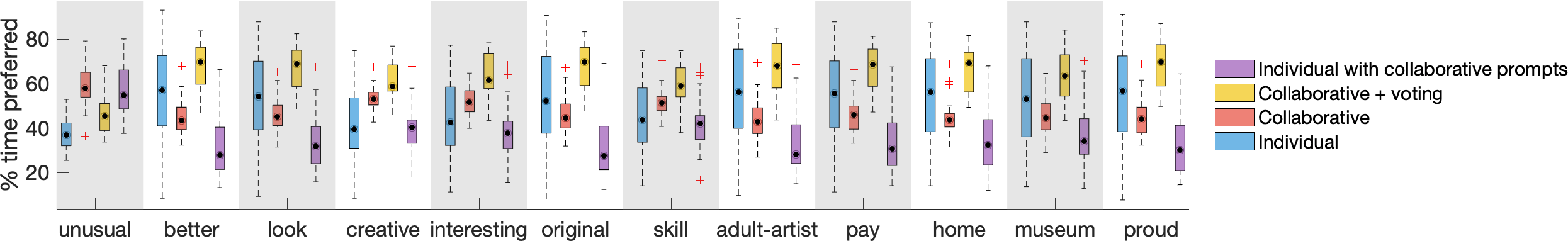}
    \caption{\textbf{Collaborative + voting} sketches are consistently preferred by human subjects over sketches from other scenarios across a variety of dimensions, and notably are rated as most creative. Notice the high variance in \textbf{Individual} sketches.}
    \label{fig:results}
\end{figure*}

\section{Evaluation}

Example sketches from these scenarios are shown in Figure~\ref{fig:example_sketches}\iftoggle{arxiv}{ \arxiv{as well as in Figures~\ref{fig:all_ind}, \ref{fig:all_col}, \ref{fig:all_col_vot}, \ref{fig:all_ind_prmpt}}}{}. Before we discuss properties of the final sketch, it is worth considering the evolution of a sketch as it is being created. \textbf{Collaborative} sketches evolve in several interesting ways: what seems like the main subject of a sketch changes in a few iterations (Figure~\ref{fig:evolution}, left), given seemingly incoherent strokes, subsequent subjects try and emphasize regions that could lead to meaningful structures in the sketch for future subjects to build on (Figure~\ref{fig:evolution}, right), and subjects use the color white or other strategies to try and cover parts of the sketch they think are contributing negatively to it. \iftoggle{arxiv}{\arxiv{More examples of sketches evolving across iterations can be found in Figures~\ref{fig:evo_col} and \ref{fig:evo_col_vot}.}}{}

To assess the qualitative differences between sketches produced from the 4 scenarios, we created a collage of 20 sketches from each scenario (at 20 iterations for \textbf{Collaborative} and \textbf{Collaborative + voting}, 30 for the rest). We showed pairs of collages to subjects on Amazon Mechanical Turk and asked them to describe differences that stood out. Snippets from subjects' responses are shown in Figure~\ref{fig:example_sketches}.

For a quantitative evaluation, we showed subjects pairs of sketches from two different scenarios ($i$,$j$). Each subject picked which sketch they prefer along 12 axes. Every pair was evaluated by 5 subjects resulting in 144,000 assessments: 20 (sketches from scenario $i$) $\times$ 20 (sketches from scenario $j$) $\times$ 6 (pairs of scenarios) $\times$ 12 (axes) $\times$ 5 (subjects per sketch-pair). The 12 axes were: which painting (1) seems more strange / unusual / different than typical paintings? (2) is a better painting? (3) do you like looking at more? (4) is more creative? (5) is more interesting? (6) is more original? (7) took more skill? (8) is made by an artist more likely to be an adult? (9) would you pay more for? (10) are you more likely to put up in your home? (11) is more likely to be in an art museum? (12) would you be more proud to have made yourself? Some of these axes (e.g., originality, novelty, skill) are from~\cite{velde2015}. 

The \% of times each scenario was picked over a competing scenario is shown in Figure~\ref{fig:results}. For 11 of 12 axes, including creativity, \textbf{Collaborative + voting} is preferred. \textbf{Collaborative + voting} scores well for both novelty (unusual) and quality (better, look), which we hypothesize increases its perceived creativity. \textbf{Individual} is rated well for quality but scores poorly on novelty. Across 11 axes, \textbf{Individual} has high variance due to differences in skill + motivation of individuals creating the sketches. \textbf{Collaborative} scores well on novelty, but worse on quality. \textbf{Individual with collaborative prompts} does poorly across all axes except for unusual\iftoggle{arxiv}{\arxiv{, which is visually apparent in Figure \ref{fig:all_ind_prmpt}}}{}. Of all scenarios, \textbf{Collaborative + voting} falls in the sweet spot for maximizing creativity (value + novelty).

\section{Discussion}

In what way may a machine best contribute to the collaborative creation of a sketch? It is often the case that humans may not be good at generating strokes, but can tell if a sketch looks good or not. This may suggest using machines to generate candidate strokes and having humans vote on which versions should proceed next. The machine may also contribute in a manner similar to the humans in our fourth scenario, i.e., the machine could generate textual prompts as a human draws a sketch. The prompter can have different ``personalities'' based on whether it is trained on sketches generated from \textbf{Individual} (coherent), \textbf{Collaborative} (rich but chaotic) or \textbf{Collaborative + voting} (rich with subtle details and coherent). Humans and machine can generate strokes as a team, either in co-painting scenarios as in~\cite{Cabannes2019}, or where the machine provides some visual guidance as in~\cite{shadowdraw} or via suggestions for where to draw, what colors to use, etc. as explored in \cite{Oh2018}. We can also train a machine to be a discriminator: given a few different stokes from a human, select which stroke should be added to the sketch next. 

All our sketches started with a blank canvas. We could instead start sketches with a prompt (subject of the sketch, adjective describing a desired property of the sketch, a picture to be used as inspiration for the sketch, etc.), and have this prompt persist across iterations (or not). 

It is interesting to consider ideas of ownership in the context of crowd co-creation. While no one individual may feel a complete sense of ownership of the final piece, crowd collaboration may lead to a sense of community and the satisfaction of contributing to a common cause. Finally, while our motivation was human-machine co-creation, studying human-human collaboration in general is, obviously, important and interesting in and of itself. Collaborative creative endeavors may be a fertile ground for such explorations.


\bibliographystyle{iccc}
\bibliography{iccc}

\iftoggle{arxiv}{


\begin{figure*}[h!]
    \centering
    \includegraphics[width=\textwidth]{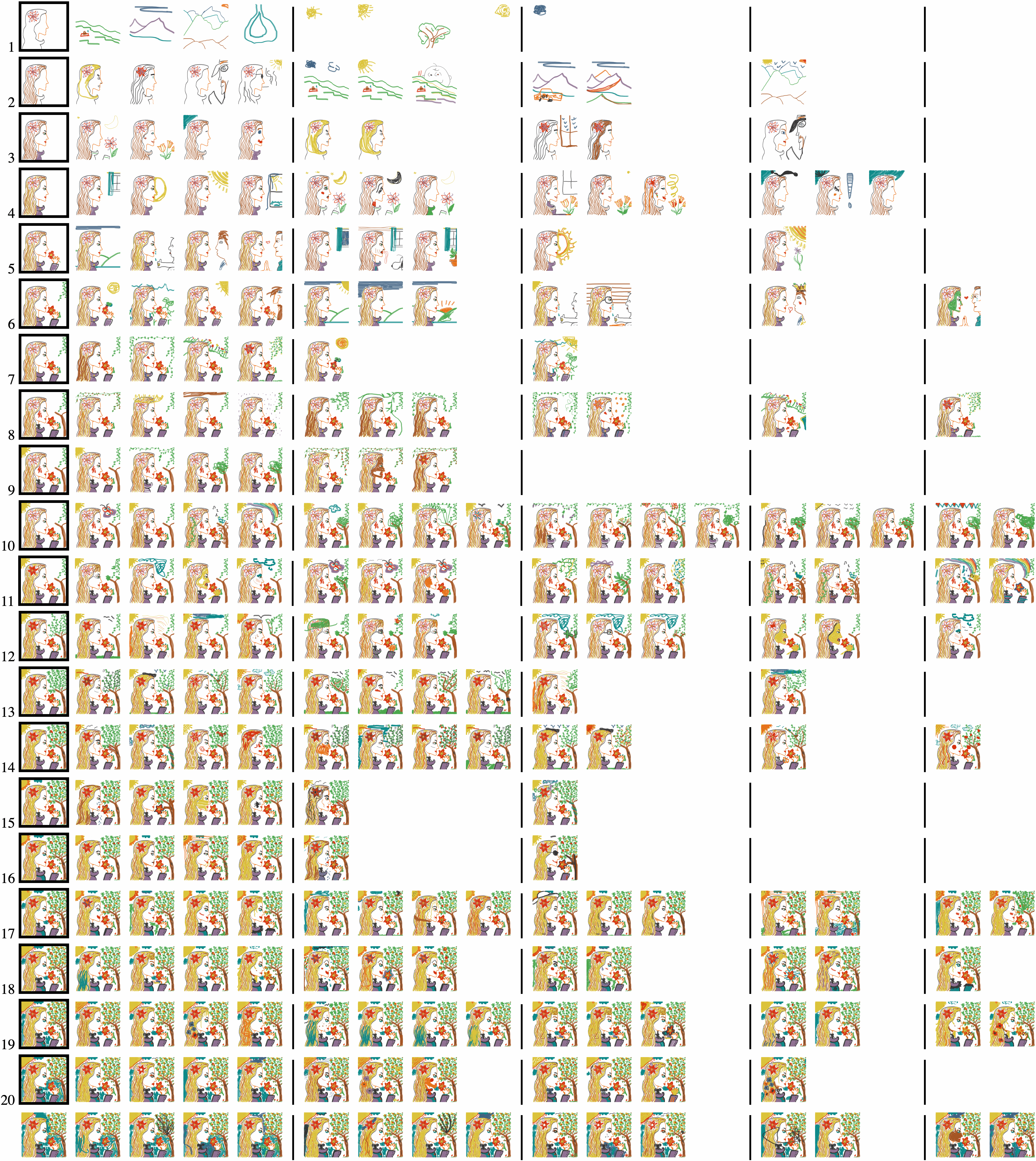}
    \caption{\arxiv{Iterations of a sketch being created in the \textbf{Collaborative + voting} scenario. Rows correspond to iterations. Each meta-column (separated by black vertical lines) shows children of the same parent. When a meta-column has five children, it's corresponding parent (outlined in black in the previous row) is selected as the next iteration of the sketch, and the five children become the next iteration of five parents. Columns have been sorted based on number of children for clarity. See Figure~\ref{fig:tree} for a clearer visualization for a few iterations. Temporal visualization: \url{https://youtu.be/JQmGALAhhMU}.}
    }
    \label{fig:whole_tree_6}
\end{figure*}

\begin{figure*}[h!]
    \centering
    \includegraphics[width=\textwidth]{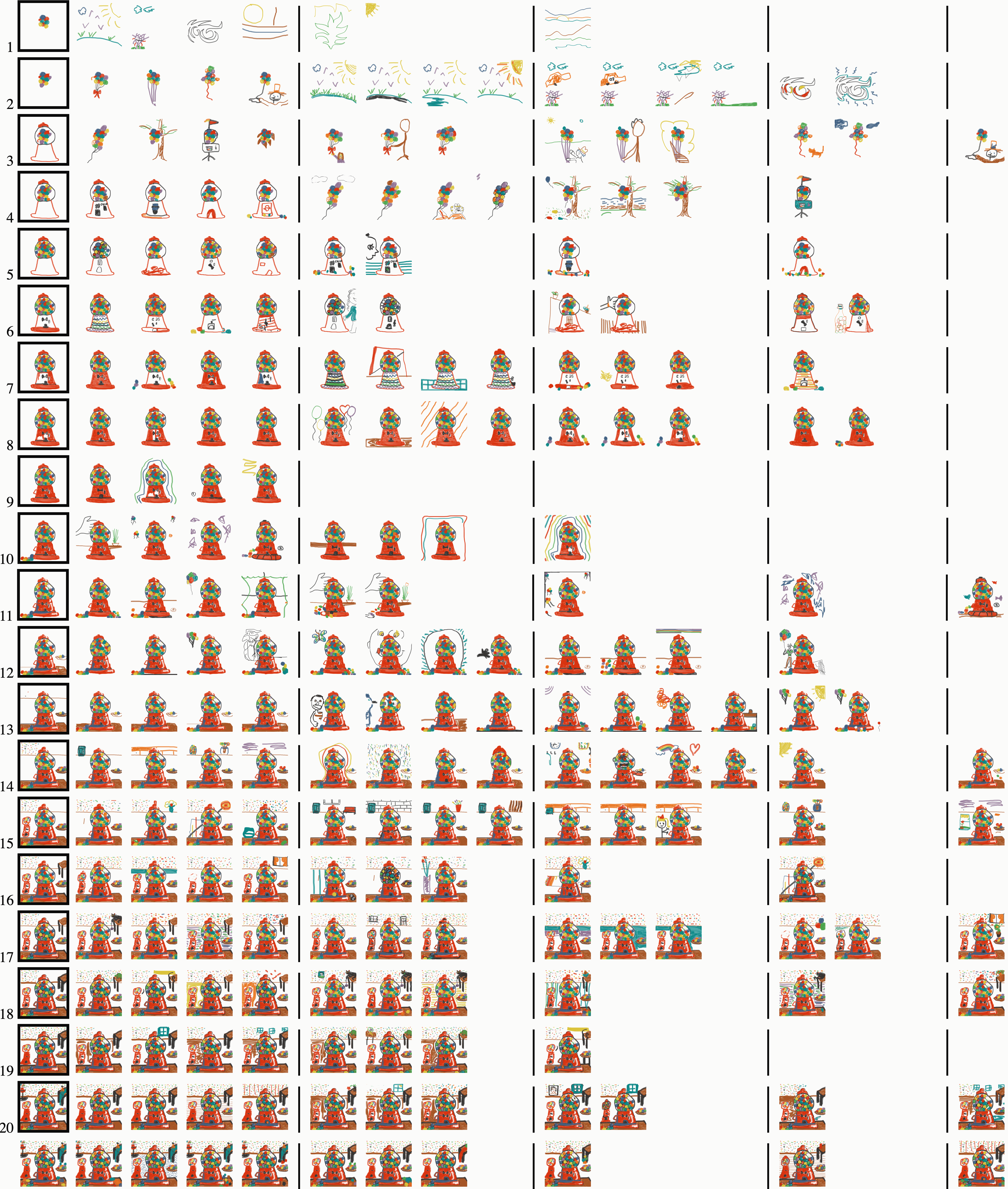}
    \caption{\arxiv{Iterations of a sketch being created in the \textbf{Collaborative + voting} scenario. Rows correspond to iterations. Each meta-column (separated by black vertical lines) shows children of the same parent. When a meta-column has five children, it's corresponding parent (outlined in black in the previous row) is selected as the next iteration of the sketch, and the five children become the next iteration of five parents. Columns have been sorted based on number of children for clarity. See Figure~\ref{fig:tree} for a clearer visualization for a few iterations. Temporal visualization: \url{https://youtu.be/JQmGALAhhMU}.}
    }
    \label{fig:whole_tree_17}
\end{figure*}

\begin{figure*}[h!]
    \centering
    \includegraphics[width=\textwidth]{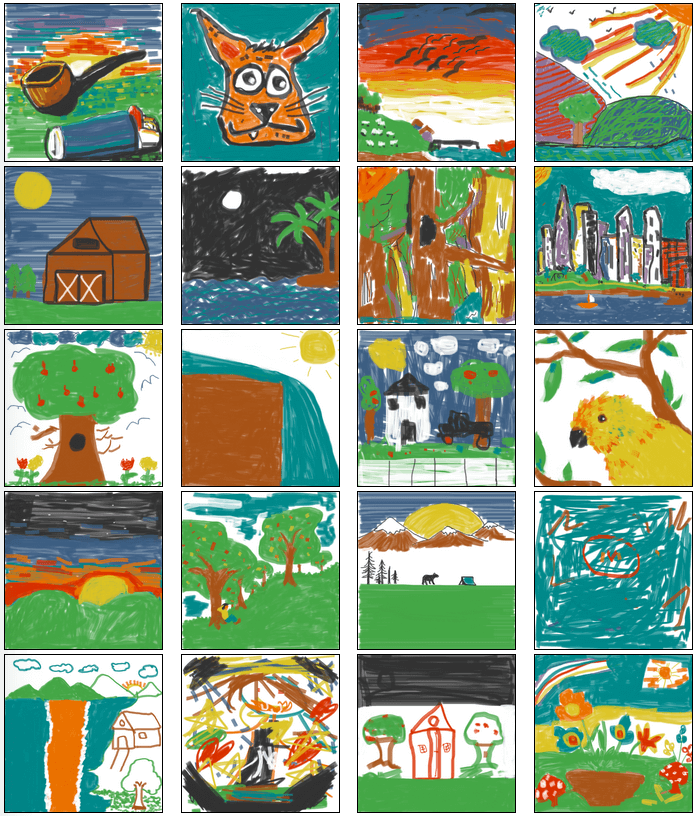}
    \caption{\arxiv{Example sketches from the \textbf{Individual} scenario. Each sketch was created by a single individual over 30 iterations.}}
    \label{fig:all_ind}
\end{figure*}

\begin{figure*}[h!]
    \centering
    \includegraphics[width=\textwidth]{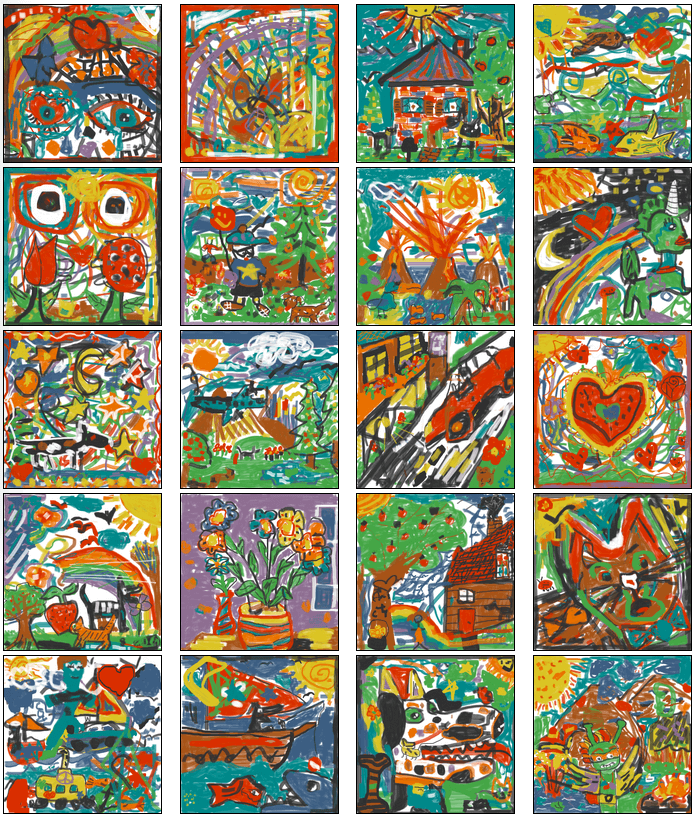}
    \caption{\arxiv{Example sketches from the \textbf{Collaborative} scenario. Each sketch was created by 30 individuals over 30 iterations.}}
    \label{fig:all_col}
\end{figure*}

\begin{figure*}[h!]
    \centering
    \includegraphics[width=\textwidth]{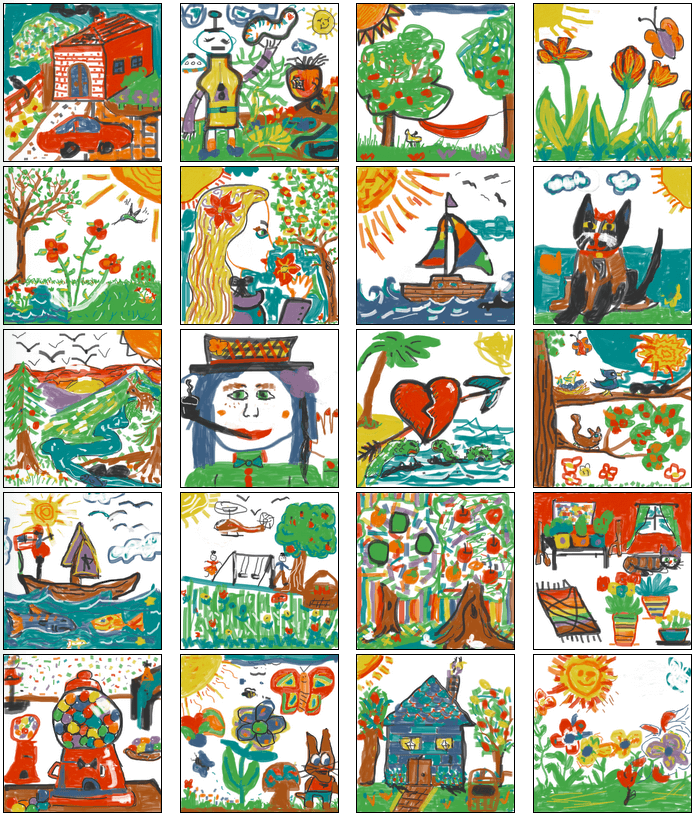}
    \caption{\arxiv{Example sketches from the \textbf{Collaborative + voting} scenario. Each sketch was created on average by 250 individuals over 20 iterations. See text for details.}}
    \label{fig:all_col_vot}
\end{figure*}

\begin{figure*}[h!]
    \centering
    \includegraphics[width=\textwidth]{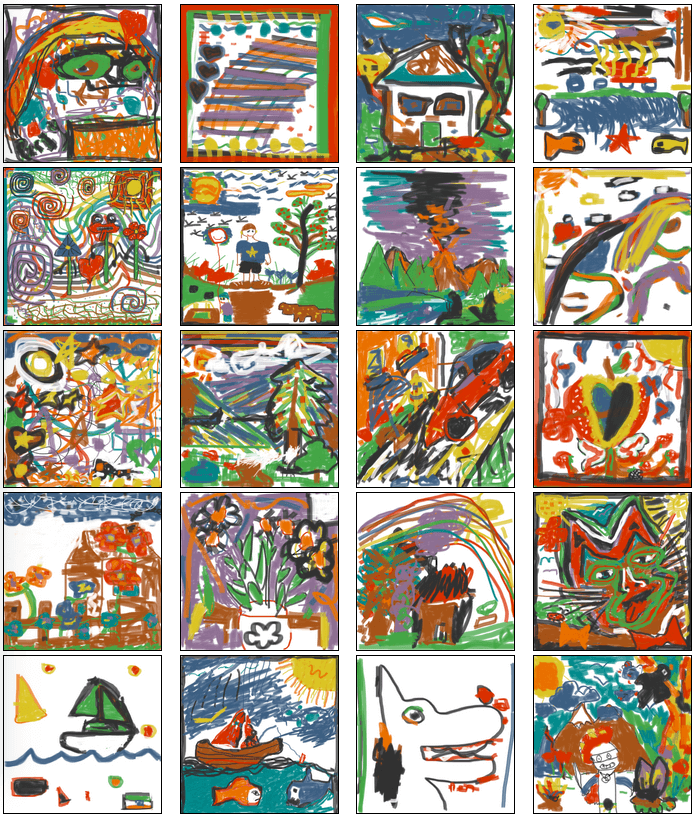}
    \caption{\arxiv{Example sketches from the \textbf{Individual with collaborative prompts} scenario. Each sketch was created by 30 individuals following text prompts. The text prompts described how sketches from the \textbf{Collaborative} scenario changed from one iteration to the next. See text for details.}}
    \label{fig:all_ind_prmpt}
\end{figure*}

\begin{figure*}[h!]
    \centering
    \includegraphics[width=\textwidth]{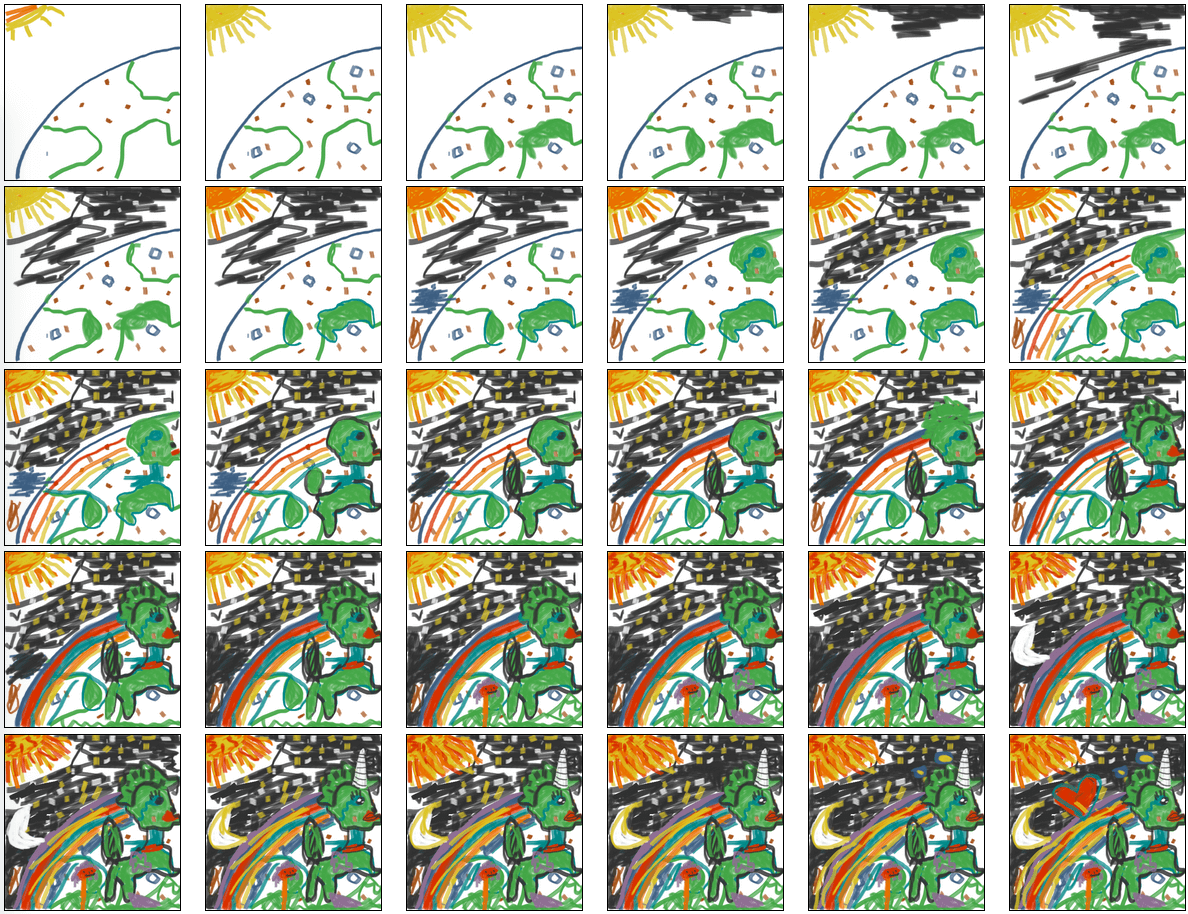}
    \caption{\arxiv{A sketch being iteratively created in the \textbf{Collaborative} scenario. Left to right, top to bottom.}}
    \label{fig:evo_col}
\end{figure*}

\begin{figure*}[h!]
    \centering
    \includegraphics[width=\textwidth]{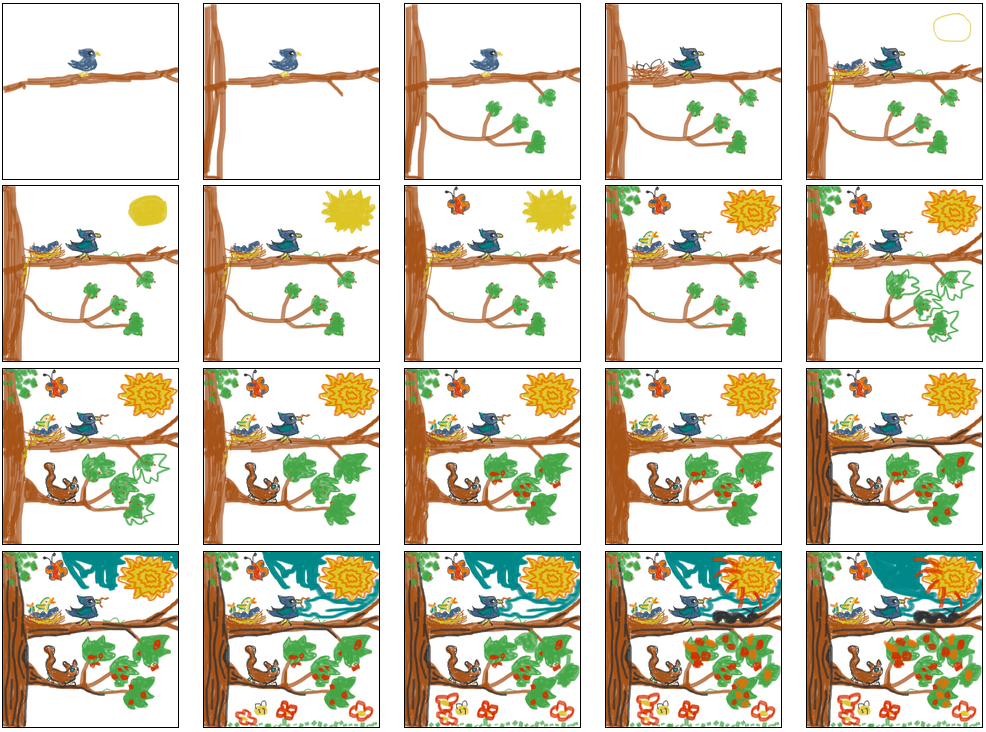}
    \caption{\arxiv{A sketch being iteratively created in the \textbf{Collaborative + voting} scenario.  Left to right, top to bottom.}}
    \label{fig:evo_col_vot}
\end{figure*}

}{}

\end{document}